\documentclass[11pt]{article}
\usepackage{colacl}
\usepackage{epsfig}
\newcommand{\epsfscaledbox}[2]{\centerline{\psfig{figure=#1,width=#2}}}
\newcommand{\set}[1]{\{#1\}}

\newcommand{\eqpunc}[1]{{\makebox[0pt][l]{$\:$\rm{#1}}}}
\newcommand{\dwa}{distance-weighted averaging}

\newcommand{\JS}{JS}
\newcommand{\n}{n}
\newcommand{\m}{m}
\newcommand{\prob}{P}
\newcommand{\modelprob}{\hat{\prob}}
\newcommand{\vb}{v}
\newcommand{\q}{q}
\newcommand{\qv}{\q(\vb)}
\renewcommand{\r}{r}
\newcommand{\rv}{\r(\vb)}
\newcommand{\conf}{{\rm conf}}
\newcommand{\jac}{{\rm Jac}}
\newcommand{\avg}{{\rm avg}_{\q,\r}}
\newcommand{\avgv}{\avg(\vb)}
\newcommand{\N}{N}
\newcommand{\V}{V}

\newcommand{\qsup}{\V_\q}
\newcommand{\rsup}{\V_\r}

\newcommand{\both}{\V_{\q\r}}

\newcommand{\leftbar}{\left\vert}
\newcommand{\rightbar}{\right\vert}
\newcommand{\sizeof}[1]{\leftbar #1 \rightbar}
\newcommand{\mysum}[1]{\mbox{$\sum\limits_{#1}$}}

\newcommand{\omtwds}[1]{#1}

\author{Lillian Lee \\ Department of Computer Science \\ Cornell
University \\ Ithaca, NY 14853-7501 \\ {\tt llee@cs.cornell.edu}}
\title{\vspace{-65pt}
{\normalsize \tt \hfill A slightly different version appeared in {\em Proceedings of the
37th ACL},  1999} \\ \mbox{} \\
Measures of Distributional Similarity}
\begin{document}
\maketitle

\begin{abstract}We study distributional similarity measures for
the purpose of improving probability estimation for unseen cooccurrences.
Our contributions are three-fold: an empirical comparison of a broad
range of measures; a classification of similarity functions based on the information
that they incorporate; and the introduction of a novel function that is
superior at evaluating potential proxy distributions.
\end{abstract}

\bibliographystyle{acl}

\section{Introduction}

An inherent problem for statistical methods in natural language
processing is that of sparse data --- the inaccurate representation in
any training corpus of the probability of low frequency events.  In
particular, reasonable events that happen to not occur in the training
set may mistakenly be assigned a probability of zero.  These {\em
unseen} events generally make up a substantial portion of novel 
data; for example, \newcite{Essen+Steinbiss:92a} report that 12\% of
the test-set bigrams in a 75\%-25\% split of one million words did not
occur in the training partition.

We consider here the question of how to estimate the conditional
cooccurrence probability
$P(\vb|\n)$ of an unseen word pair $(\n,\vb)$ drawn from some finite set
$N \times V$.
Two state-of-the-art technologies are Katz's \shortcite{Katz:87a} {\em backoff} method
and Jelinek and Mercer's \shortcite{Jelinek+Mercer:80a} interpolation
method.  Both use $P(\vb)$ to
estimate $P(\vb|\n)$ when $(\n,\vb)$ is unseen, essentially
ignoring the identity of $\n$.  

An alternative approach is {\em \dwa}, which arrives at an estimate for unseen cooccurrences by combining estimates for cooccurrences
involving similar words:\footnote{The term
``similarity-based'', which we have used previously, has been
applied to describe other  models as well \nocite{Karov+Edelman:98a} \nocite{Lee:thesis} (L. Lee, 1997; Karov and Edelman,
1998).}
\begin{equation}
\modelprob(\vb|\n) = \frac{\sum_{\m \in {\cal S}(\n)} {\rm sim}(\n,\m)
\prob(\vb|\m)}{\sum_{\m \in {\cal S}(\n)}{\rm sim}(\n,\m)} \,,
\label{eq:dwa}
\end{equation}
where ${\cal S}(\n)$ is a set of candidate similar words and ${\rm
sim}(\n, \m)$ is a function of the similarity between $n$ and $\m$.
We focus on {\em distributional} rather than semantic
similarity (e.g., \newcite{Resnik:95b})
because the goal of \dwa\ is to smooth probability distributions ---
although the words ``chance'' and ``probability'' are synonyms, the
former may not be a good model for predicting what cooccurrences the
latter is likely to participate in.  

There are many plausible measures of distributional similarity.
In previous work \cite{Dagan+Lee+Pereira:99a}, we compared the
performance of three different functions: the Jensen-Shannon divergence (total
divergence to the average),
the $L_1$ norm, and the confusion probability.  Our experiments on a
frequency-controlled pseudoword disambiguation task showed that using
any of the three in a \dwa\ scheme yielded large  improvements over
Katz's backoff smoothing method in predicting unseen coocurrences.
Furthermore, by using a restricted version of model (\ref{eq:dwa})
that stripped incomparable parameters, we were able to
empirically demonstrate that the confusion probability is
fundamentally worse at selecting useful similar words.  D. Lin also found that the choice of similarity function
can affect the quality of automatically-constructed thesauri to a
statistically significant degree \shortcite{Lin:98b} and the ability
to determine common morphological roots by as much as 49\% in
precision \shortcite{Lin:98a}.

These empirical results indicate that investigating different
similarity measures can lead to improved natural language processing.
On the other hand, while there have been many similarity measures
proposed and analyzed in the information retrieval literature
\cite{Jones+Furnas:87a}, there has been some doubt expressed in that
community that the choice of similarity metric has any practical
impact:
\vspace*{-.05in}
\begin{quote} Several
authors have pointed out that the difference in retrieval performance
achieved by different measures of association is insignificant,
providing that these are appropriately normalised.
\cite[pg. 38]{Rijsbergen:79a}
\end{quote}  \vspace*{-.05in} But no contradiction arises because, as van Rijsbergen
continues, ``one would expect this since most measures incorporate the
same information''.  In the language-modeling domain, there is
currently no agreed-upon best similarity metric because there is no
agreement on what the ``same information'' --- the key data that a
similarity function should incorporate --- is.

The overall goal of the work described here was to discover these key
characteristics.  To this end, we first compared a number of common
similarity measures, evaluating them in a parameter-free way on a
decision task.  When grouped by average performance, they fell into
several coherent classes, which corresponded to the extent to which
the functions focused on the intersection of the {\em supports}
(regions of positive probability) of the distributions.  Using this
insight, we developed an information-theoretic metric, the {\em skew
divergence}, which incorporates the support-intersection data in an
asymmetric fashion.  This function yielded the best performance
overall: an average error rate reduction of 4\% (significant at the
.01 level) with respect to the Jensen-Shannon divergence, the best
predictor of unseen events in our earlier experiments
\cite{Dagan+Lee+Pereira:99a}.

Our contributions are thus three-fold: an empirical comparison of a
broad range of similarity metrics using an evaluation methodology that
factors out inessential degrees of freedom; a proposal, building on
this comparison, of a characteristic for classifying similarity
functions; and the introduction of a new similarity metric
incorporating this characteristic that is superior at evaluating
potential proxy distributions.

\section{Distributional Similarity Functions}
\label{sec:fns}

In this section, we describe the seven distributional similarity
functions we initally evaluated.\footnote{Strictly speaking, some of
these functions are dissimilarity measures, but each such function $f$
can be recast as a similarity function via the simple transformation
$C - f$, where $C$ is an appropriate constant.  Whether we mean $f$ or
$C - f$ should be clear from context.}
For concreteness, we choose $\N$ and $\V$ to be the set of nouns and
the set of transitive verbs, respectively; a cooccurrence pair
$(\n,\vb)$ results when $\n$ appears as the head noun of the direct
object of $\vb$.  We use $\prob$ to denote probabilities assigned by a
base language model (in our experiments, we simply used unsmoothed
relative frequencies derived from training corpus counts).

Let $\n$ and $\m$ be two nouns whose distributional similarity is to
be determined; for notational simplicity, we write $\qv$ for
$\prob(\vb|\n)$ and $\rv$ for $\prob(\vb|\m)$, their respective
conditional verb cooccurrence probabilities.

\begin{figure*}
\begin{eqnarray*}
\mbox{Euclidean distance} \hspace*{.2in}  L_2(\q,\r) & = & \sqrt{\sum_\vb \left(\qv - \rv\right)^2}
\\
\mbox{$L_1$ norm} \hspace*{.2in} L_1(\q,\r)& =& \sum_\vb |\qv - \rv| \\
\mbox{cosine} \hspace*{.2in}  \cos(\q,\r) & = &  \frac{\sum_\vb
\qv \rv}{\sqrt{\sum_\vb \qv^2}  \sqrt{\sum_\vb \rv^2}} \\
\mbox{Jaccard's coefficient} \hspace*{.2in}  \jac(\q,\r)  & = &
\frac{\sizeof{ \{\vb\: : \: \qv > 0 \mbox{ and } \rv > 0
\}}}{\sizeof{\{ \vb \: | \: \qv > 0 \mbox{ or } \rv > 0 \}}}
\end{eqnarray*}
\caption{\label{fig:famous} Well-known functions}
\end{figure*}
Figure \ref{fig:famous} lists several familiar functions.  The cosine metric and Jaccard's
coefficient are commonly used in information retrieval as measures of
association \cite{Salton+McGill:83a}.  Note that
Jaccard's coefficient differs from all the other measures we consider
in that it is essentially {\em combinatorial}, being based only on the
sizes of the supports of $\q$, $\r$, and $\q\cdot\r$ rather than the
actual values of the distributions.

Previously, we found the {\em Jensen-Shannon di\-ver\-gence}
\nocite{JLin:91} \nocite{Rao:82a} (Rao, 1982; J. Lin, 1991) to be a
useful measure of the distance between distributions:
\omtwds{\begin{displaymath} \nonumber \JS(\q,\r) = \frac{1}{2} \left[
D\left(\q \, \Big \Vert \, \avg \right) + D\left( \r \, \Big \Vert \,
\avg \right) \right] \eqpunc{.}
\end{displaymath}}
The function $D$ is the {\em KL divergence}, which measures the (always
nonnegative) average inefficiency in using one distribution to code for
another \cite{Cover+Thomas:91a}:
\begin{displaymath}
D(p_1(V) \, \Vert \, p_2(V)) = \sum_v p_1(v) \log \frac{p_1(v)}{p_2(v)}
\eqpunc{.}
\end{displaymath}
The function $\avg$ denotes the average
distribution $\avgv = (\qv + \rv)/2$; observe that its use ensures
that the Jensen-Shannon divergence is always defined. In contrast, $D(\q || \r)$ is
undefined if $\q$ is not absolutely continuous with respect to $\r$
(i.e., the support of $\q$ is not a subset of the support of $\r$).

  The {\em confusion probability}
has been used by several authors to smooth word cooccurrence
probabilities
\cite{Sugawara+al:85a,Essen+Steinbiss:92a,Grishman+Sterling:93a}; it
measures the degree to which word $\m$ can be substituted into the
contexts in which $\n$ appears.  If the base language model
probabilities obey certain Bayesian consistency conditions
\cite{Dagan+Lee+Pereira:99a}, as is the case for relative frequencies,
then we may write the confusion probability as follows:
\omtwds{\begin{displaymath} \conf(\q,\r,\prob(\m)) = \sum_\vb \qv \rv
\frac{\prob(\m)}{\prob(\vb)} \eqpunc .
\end{displaymath}} Note that it
incorporates unigram probabilities as well as the two distributions
$\q$ and $\r$.  

Finally, 
{\em Kendall's $\tau$}, which  
appears in work
on clustering similar adjectives
\nocite{Hatzivassiloglou+McKeown:93a,Hatzivassiloglou:96a} 
(Hatzi\-vassi\-lo\-glou and McKeown, 1993; Hatzi\-vassi\-lo\-glou, 1996), is a
nonparametric measure of the association between random variables
\cite{Gibbons:93a}.  In our context, it looks for
correlation between the behavior of $\q$ and $\r$ on pairs of verbs.  Three
versions exist; we use the simplest, $\tau_a$, here:
\omtwds{\begin{displaymath} \tau(\q,\r) = \sum_{\vb_1,\vb_2} \frac{{\rm
sign}\left[(\q(\vb_1) - \q(\vb_2)) (\r(\vb_1) - \r(\vb_2))\right]} {
{2{{|V|} \choose 2}}},
\end{displaymath}}
where ${\rm sign}(x)$ is 1 for positive arguments, $-1$ for negative
arguments, and 0 at 0. The intuition behind Kendall's $\tau$ is as
follows. Assume all verbs have distinct conditional probabilities.  If
sorting the verbs by the likelihoods assigned by $\q$ yields exactly
the same ordering as that which results from ranking them according to
$\r$, then $\tau(\q,\r) = 1$; if it yields exactly the opposite
ordering, then $\tau(\q ,\r) = -1$. We treat a value of $-1$ as
indicating extreme dissimilarity.\footnote{Zero would also be a
reasonable choice, since it indicates zero correlation between $\q$
and $\r$.  However, it would then not be clear how to average in the
estimates of negatively correlated words in equation (\ref{eq:dwa}).}

It is worth noting at this point that there are several well-known
measures from the NLP literature that we have omitted from our
experiments.  Arguably the most widely used
is the {\em mutual information} (Hindle, 1990; Church and Hanks, 1990;
Dagan et al., 1995; Luk, 1995; D. Lin, 1998a).
\nocite{Hindle:90a,Church+Hanks:90a,Dagan+Marcus+Markovitch:95a,Luk:95a}
It does not apply in the present setting because it does not measure
the similarity between two arbitrary probability
distributions (in our case, $P(V|n)$ and $P(V|m)$), but rather the similarity
between a joint distribution $P(X_1,X_2)$ and the corresponding product
distribution $P(X_1)P(X_2)$.  Hamming-type metrics
\cite{Cardie:93a,Zavrel+Daelemans:97a} are intended for data with
symbolic features, since they count feature label mismatches, whereas
we are dealing feature values that are probabilities.  Variations of
the {\em value difference metric} \cite{Stanfill+Waltz:86a} have been
employed for supervised disambiguation (Ng and H.B. Lee, 1996; Ng,
1997); \nocite{Ng+Lee:96a,Ng:97b} but it is not reasonable in language
modeling to expect training data tagged with correct probabilities.
The {\em Dice coefficient}
\nocite{Smadja+McKeown+Hatzivassiloglou:96a}
 (Smadja et al., 1996;
D. Lin, 1998a, 1998b)
\cite{Kay+Roscheisen:93a}
 is monotonic in Jaccard's coefficient
\cite{Rijsbergen:79a}, so its inclusion in our experiments would be
redundant.  Finally, we did not use the KL divergence because it
requires a smoothed base language model.

\section{Empirical Comparison}
\label{sec:eval}
We evaluated the similarity functions introduced in the previous
section on a binary dec\-ision task, using the same experimental
framework as in our previous preliminary comparison
\cite{Dagan+Lee+Pereira:99a}.  That is, the data consisted of the
verb-object cooccurrence pairs in the 1988 Associated Press newswire
involving the 1000 most frequent nouns, extracted via Church's
\shortcite{Church:88a} and Yarowsky's processing tools.  587,833
(80\%) of the pairs served as a training set from which to calculate
base probabilities.  From the other 20\%, we
prepared test sets as follows: after
discarding pairs occurring in the training data (after all, the point
of similarity-based estimation is to deal with unseen pairs), we split
the remaining pairs into five partitions, and replaced each noun-verb
pair $(\n, \vb_1)$ with a noun-verb-verb triple
$(\n, \vb_1, \vb_2)$ such that $P(\vb_2) \approx P(\vb_1)$. The task
for the language model under evaluation was to reconstruct which of
$(\n, \vb_1)$ and $(\n,\vb_2)$ was the original cooccurrence.
Note that by construction, $(\n,\vb_1)$ was always the correct answer,
and furthermore, methods relying solely on unigram frequencies would
perform no better than chance.  Test-set performance was measured by
the error rate, defined as
\[
\frac{1}{T} (\mbox{\# of incorrect choices } + (\mbox{\# of
ties})/2) \eqpunc,
\]
where $T$ is the number of test triple tokens in the set, and a tie
results when both alternatives are deemed equally likely by the
language model in question.

To perform the evaluation, we incorporated each similarity function
into a simple decision rule as follows. As above, let
$(n,\vb_1,\vb_2)$ be a test instance. For a given similarity
measure $f$ and neighborhood size $k$, let ${\cal S}_{f,k}(\n)$ denote
the $k$ most similar words to $\n$ according to $f$. We define the
{\em evidence} $E_{f,k}(\n,\vb_1)$ for $v_1$ as the number of
neighbors $m \in {\cal S}_{f,k}(\n)$ such that
$P(\vb_1|m)>P(\vb_2|m)$; similarly, the evidence for $\vb_2$ is the
number of the $k$ closest neighbors that favor $\vb_2$ over $\vb_1$.
Then, the decision rule is to choose the verb alternative with the
greatest evidence.

The reason we used a restricted version of the \dwa\ model was that we
sought to discover fundamental differences in behavior. Because we
have a binary decision task, $E_{f,k}(\n,\vb_1)$ simply counts the
number of $k$ nearest neighbors to $\n$ that make the right
decision.  If we have two functions $f$ and $g$ such that $E_{f,k}(\n,\vb_1) >
E_{g,k}(\n,\vb_1)$, then the $k$ most similar words according to $f$
are on the whole better predictors than the $k$ most similar words
according to $g$; hence, $f$ induces an inherently better similarity
ranking for \dwa.  The difficulty with using the full model (Equation
(\ref{eq:dwa})) for comparison purposes is that fundamental
differences can be obscured by issues of weighting.  For example,
suppose the probability estimate $\sum_\vb (2
-L_1(\q,\r)) \cdot \rv$ (suitably normalized) performed poorly.  We would
not be able to tell whether the cause was an
inherent deficiency in the $L_1$ norm or just a poor choice of weight
function --- perhaps $(2-L_1(\q,\r))^2$ would have yielded better estimates.

Figure \ref{fig:base} shows how the average error rate varies with $k$
for the seven similarity metrics introduced above.  As previously
mentioned, a steeper slope indicates a better similarity ranking.
\omtwds{
\begin{figure*}[htb]
\epsfscaledbox{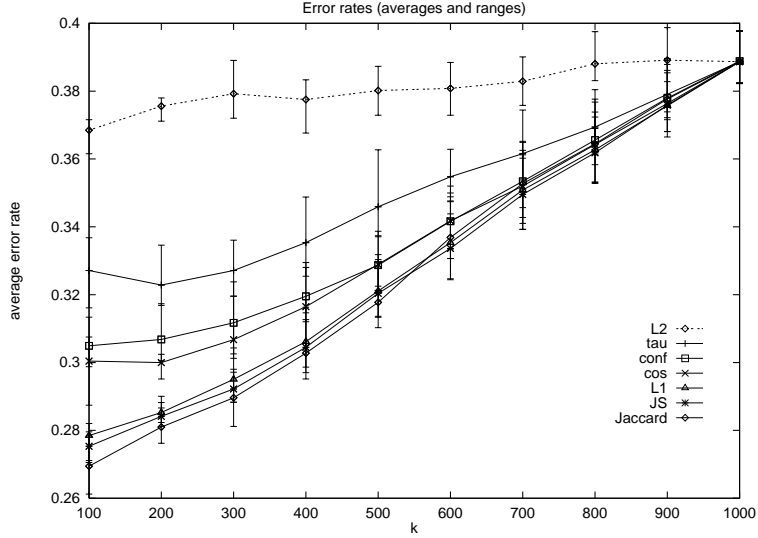}{4in}
\caption{\label{fig:base}  Similarity metric performance. Errorbars denote
the range of error rates over the five test sets.  Backoff's average
error rate was 51\%.}
\end{figure*}
} 

All the curves have a generally upward trend but always lie far below
backoff (51\% error rate). They meet at $k=1000$ because ${\cal
S}_{f,1000}(\n)$ is always the set of all nouns.  We see that the
functions fall into four groups: (1) the $L_2$ norm; (2) Kendall's
$\tau$; (3) the confusion probability and the cosine metric; and (4)
the $L_1$ norm, Jensen-Shannon divergence, and Jaccard's coefficient.

\begin{figure}[ht]
\epsfscaledbox{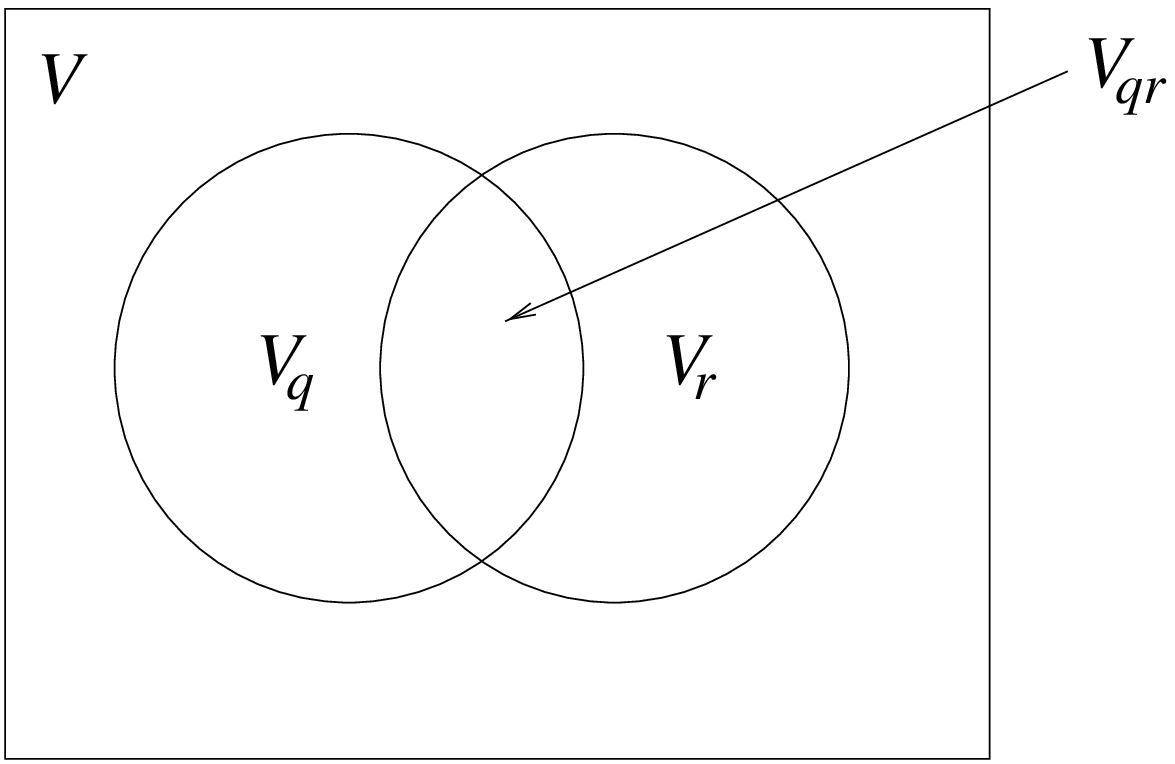}{2in}
\caption{\label{fig:supports} Supports on $\V$}
\end{figure}

We can account for the similar performance of various metrics by
analyzing how they incorporate information from the intersection of
the supports of $\q$ and $\r$.  (Recall that we are using $\q$ and
$\r$ for the conditional verb cooccurrrence probabilities of two nouns
$\n$ and $\m$.)  Consider the following supports (illustrated in
Figure \ref{fig:supports}):
\begin{eqnarray*} 
\qsup  & = & \set{\vb \in \V \: : \: \qv > 0}  \\
\rsup  & =  &\set{\vb \in \V \: : \: \rv > 0} \\
\both   &=  & \set{\vb \in \V \: : \: \qv \rv  > 0}  = \qsup \cap
\rsup 
\end{eqnarray*}

 % end omtwds

\noindent We can rewrite the similarity functions from Section \ref{sec:fns} in
terms of these sets, making use of the identities $\sum_{\vb \in \qsup
\setminus \both} \qv + \sum_{\vb \in \both} \qv = \sum_{\vb \in \rsup \setminus \both}
\rv + \sum_{\vb \in \both} \rv = 1$.  Table \ref{bothfns} lists these
alternative forms in order of performance.
\omtwds{
\begin{table*}[htb]
%\fbox{%
\def\arraystretch{2}
\fbox{
\begin{tabular}{ll}
$L_2(\q,\r)$ & = $\sqrt{\mysum{\qsup} \qv^2 - 2\mysum{\both}\qv\rv +
\mysum{\rsup}\rv^2}$ \\ \hline
$\tau(\q,\r) \cdot 2 {{|V|} \choose 2}$  & =  $2 \sizeof{\both}\sizeof{\V \setminus (\qsup \cup
\rsup)} -  2 \sizeof{\qsup \setminus \both}\sizeof{\rsup \setminus
\both}$ \\
      &  $ \qquad+
\mysum{\vb_1 \in (\qsup \triangle \rsup)} \: \mysum{\vb_2 \in \both}{\rm sign}[ (\q(\vb_1) - \q(\vb_2))
 (\r(\vb_1) - \r(\vb_2))]$ \\
	&  $ \qquad + \mysum{\vb_1 \in \both} \: \mysum{\vb_2 \in \qsup \cup \rsup} {\rm sign}[ (\q(\vb_1) - \q(\vb_2))
 (\r(\vb_1) - \r(\vb_2))] $ \\  \hline
$\conf(\q,\r, \prob(\m))$ & =  $\prob(\m) \mysum{\vb \in \both}
\qv\rv / \prob(\vb)$ \\
$\cos(\q,\r)$ & =  $\mysum{\vb \in \both} \qv \rv
	({\mysum{\vb \in \qsup}\qv^2 \mysum{\vb \in
\rsup}\rv^2})^{-1/2}$ \\ \hline
$L_1(\q,\r)$ & =  $2 - \mysum{\vb \in \both} (\left\vert\qv -
\rv \right\vert  - \qv -\rv)$ \\ 
$\JS(\q,\r)$ & =  $\log 2 + \frac{1}{2} \mysum{\vb \in \both} \left( h(\qv +
\rv) - h(\qv) - h(\rv) \right) \, ,  \qquad  h(x) = -x \log x $ \\
$\jac(\q,\r)$ & = $\sizeof{\both}/\sizeof{\qsup \cup \rsup}$
\end{tabular}
} % end fbox
\caption{\label{bothfns} Similarity functions, written in terms of
sums over supports and grouped by average performance.  $\setminus$
denotes set difference; $\triangle$ denotes symmetric set
difference. }
\end{table*}
} % end omtwds
We see that for the non-combinatorial functions, the groups correspond
to the degree to which the measures rely on the 
verbs in $\both$.  The Jensen-Shannon divergence and the $L_1$
norm can be computed simply by knowing the values of $\q$ and $\r$ on
$\both$.  For the cosine and the confusion probability, the
distribution values on $\both$ are key, but other information is also
incorporated.  The statistic $\tau_a$ takes into account
all verbs, including those that occur neither with $\n$ nor $\m$.
Finally, the Euclidean distance is quadratic in verbs outside $\both$;
indeed, \newcite{Kaufman+Rousseeuw:90} note that it is ``extremely
sensitive to the effect of one or more outliers'' (pg. 117).

The superior performance of $\jac(\q,\r)$ seems to underscore the
importance of the set $\both$.  Jaccard's coefficient ignores the
values of $\q$ and $\r$ on $\both$; but we see that simply knowing the
size of $\both$ relative to the supports of $\q$ and $\r$ leads to
good rankings.  

\section{The Skew Divergence}
\label{sec:skew}

Based on the results just described, it appears that it is desirable
to have a similarity function that focuses on the verbs that cooccur
with both of the nouns being compared.  However, we can make a further
observation: with the exception of the confusion probability, all the
functions we compared are symmetric, that is, $f(\q,\r) = f(\r,\q)$.
But the substitutability of one word for another need not symmetric.
For instance, ``fruit'' may be the best possible approximation to
``apple'', but the distribution of ``apple'' may not be a suitable
proxy for the distribution of ``fruit''.\footnote{On a related note,
an anonymous reviewer cited the following example from the psychology
literature: we can say Smith's lecture is like a sleeping pill, but
``not the other way round''.}

In accordance with this insight, we developed a novel asymmetric
generalization of the KL divergence, the {\em
$\alpha$-skew divergence}: \omtwds{
\begin{displaymath}
 s_\alpha(\q,\r)  =  D(\r \; \Vert \, \alpha \cdot \q + (1 -
\alpha) \cdot \r) 
\end{displaymath}
} for $0 \leq \alpha \leq 1$.  It can easily be shown that $s_\alpha$
depends only on the verbs in $\both$.  Note that at $\alpha=1$, the
skew divergence is exactly the KL divergence, and $s_{1/2}$ is twice
one of the summands of $\JS$ (note that it is still asymmetric).

We can think of $\alpha$  as a degree of confidence in the
empirical distribution $\q$; or, equivalently, $(1 - \alpha)$ can be
thought of as controlling the amount by which one smooths $\q$ by
$\r$.  Thus, we can view the skew divergence  as an
approximation to the KL divergence to be used when sparse data
problems would cause the latter measure to be undefined. 

Figure \ref{fig:skew} shows the performance of $s_\alpha$ for $\alpha
= .99$.  It performs better than all the other functions; the
difference with respect to Jaccard's coefficient is statistically
significant, according to the paired $t$-test, at all $k$ (except
$k=1000$), with significance level .01 at all $k$ except 100, 400, and
1000.  \omtwds{
\begin{figure*}[htb]
\epsfscaledbox{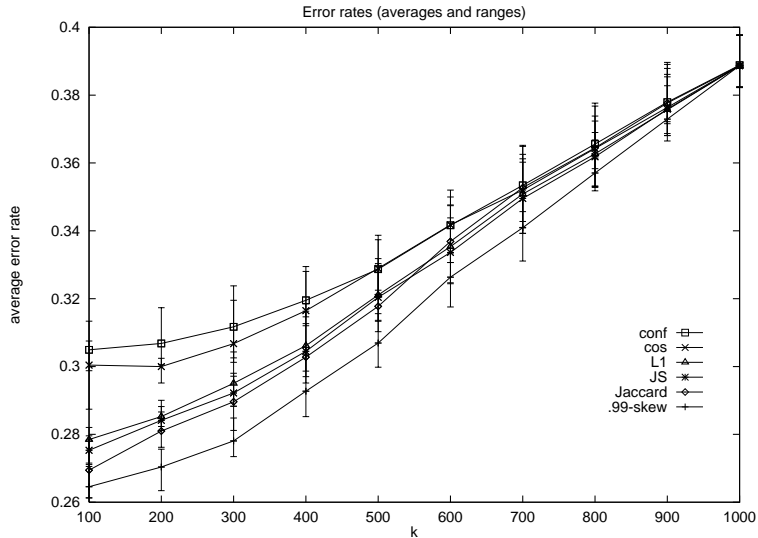}{4in}
\caption{\label{fig:skew} Performance of the skew divergence with
respect to the best functions from Figure \ref{fig:base}. }
\end{figure*}
} % end omtwds

\section{Discussion} 
\label{sec:discuss}
In this paper, we empirically evaluated a number of distributional
similarity measures, including the skew divergence, and analyzed their
information sources. \nocite{Jones+Furnas:87a} We observed that the
ability of a similarity function $f(\q,\r)$ to select useful nearest
neighbors appears to be correlated with its focus on the intersection
$\both$ of the supports of $\q$ and $\r$.  This is of interest from a
computational point of view because $\both$ tends to be a relatively
small subset of $V$, the set of all verbs.  Furthermore, it suggests
downplaying the role of negative information, which is encoded by verbs
appearing with exactly one noun, although the Jaccard
coefficient does take this type of information into account.

Our explicit division of $V$-space into various support regions has
been implicitly considered in other work.
\newcite{Smadja+McKeown+Hatzivassiloglou:96a} observe that for two potential mutual
translations $X$ and $Y$, the fact that $X$ occurs with translation
$Y$ indicates association; $X$'s
occurring with a translation other than $Y$
decreases one's belief in their association; but the absence of both
$X$ and $Y$ yields no information.  In essence, Smadja et al. argue
that information from the union of supports, rather than the just the
intersection, is important.
D. Lin \shortcite{Lin:97a,Lin:98b} takes an axiomatic approach to
determining the characteristics of a good similarity measure. Starting
with a formalization (based on certain assumptions) of the intuition
that the similarity between two events depends on both their
commonality and their differences, he derives a unique similarity
function schema.  The
definition of commonality is left to the user (several different
definitions are proposed for different tasks).  

We view the empirical approach taken in this paper as complementary to
Lin's.  That is, we are working in the context of a particular
application, and, while we have no mathematical certainty of the
importance of the ``common support'' information, we did not assume it
{\em a priori}; rather, we let the performance data guide our thinking.

Finally, we observe that the skew metric seems quite promising.  We
conjecture that appropriate values for $\alpha$ may
inversely correspond to the degree of sparseness in the data, and
intend in the future to test this conjecture on larger-scale
prediction tasks.  We also plan to evaluate skewed versions of the
Jensen-Shannon divergence proposed by \newcite{Rao:82a} and J. \newcite{JLin:91}.

\section{Acknowledgements}  
Thanks to Claire Cardie, Jon Kleinberg, David McAllester, Fernando Pereira, and Stuart
Shieber for helpful discussions, the anonymous reviewers for their
insightful comments, Fernando Pereira for access to computational
resources at AT\&T, and Stuart Shieber for the opportunity to pursue
this work at Harvard University.  This material is based upon work
supported by the National Science Foundation under Grant No.  IRI9712068.
\bibliography{sim}
\end{document}